%% file: main.tex
\documentclass[10pt,twocolumn,letterpaper]{article}

\usepackage{iccv}
\usepackage{times}
\usepackage{epsfig}
\usepackage{graphicx}
\usepackage{amsmath}
\usepackage{amssymb}
\usepackage{subcaption}

\usepackage{enumitem}
\usepackage{booktabs}
\usepackage{algorithm}
\usepackage{algpseudocode}

\usepackage[pagebackref=true,breaklinks=true,letterpaper=true,colorlinks,bookmarks=false]{hyperref}

\iccvfinalcopy 

\def\iccvPaperID{} 

\ificcvfinal\fi

\begin{document}

\title{Taming Encoder for Zero Fine-tuning Image Customization\\with Text-to-Image Diffusion Models}

\author{}
\author{Xuhui Jia$^*$\quad Yang Zhao$^*$\quad  Kelvin C.K. Chan$^*$\quad Yandong Li\quad Han Zhang\\ \quad  Boqing Gong\quad  Tingbo Hou\quad  Huisheng Wang\quad  Yu-Chuan Su \\\\
Google Research\\
{\hspace{-0.3cm}\tt\footnotesize \{xhjia,\,yzhaoeric,\,kelvinckchan,\,yandongli,\,zhanghan,\,bgong,\,tingbo,\,huishengw,\,ycsu\}@google.com}

}

\twocolumn[{%
	\renewcommand\twocolumn[1][]{#1}%
	\maketitle\thispagestyle{empty}
	\begin{center}
            \vspace{-4mm}
		\centerline{\includegraphics[width=0.99\linewidth]{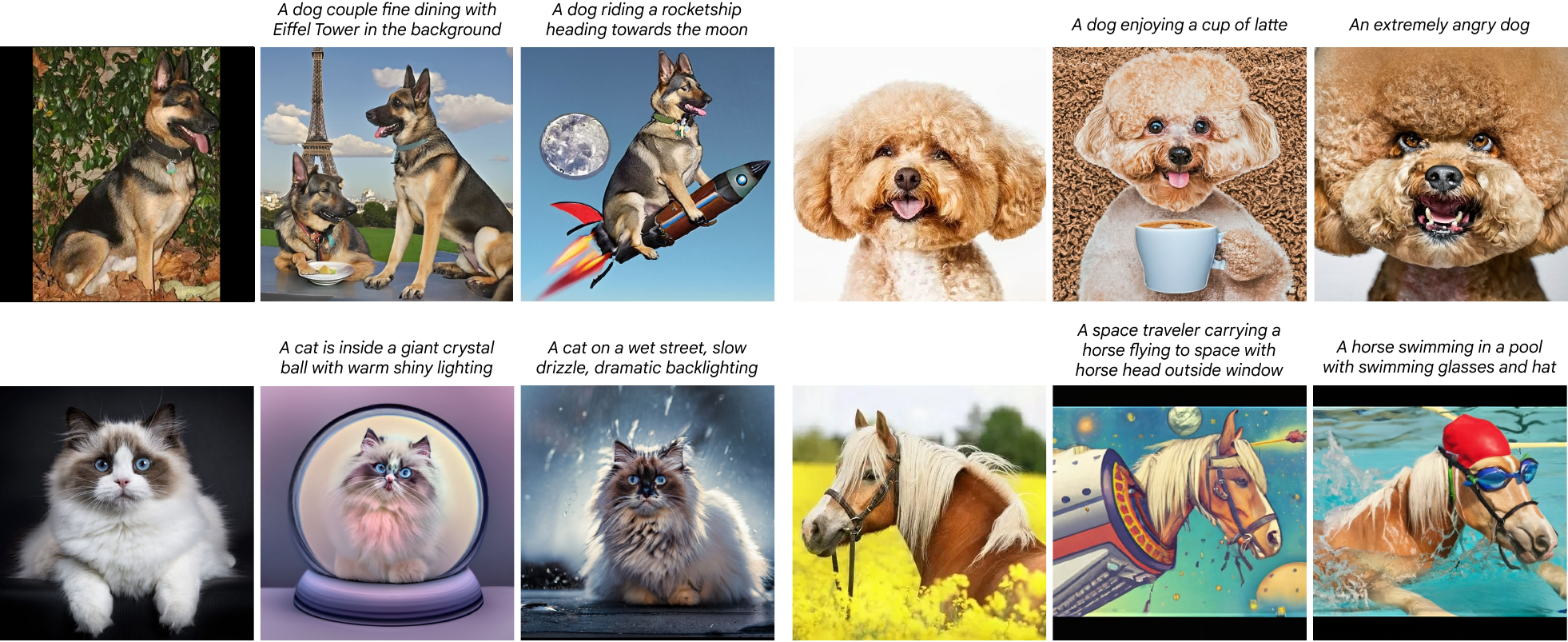}}
            \vspace{-3mm}
		\captionof{figure}{Given \textit{\textbf{one}} reference image and a text prompt, our method generates an image containing the same object seamlessly immersed with novel concepts described by the text in a \textit{\textbf{single forward pass}, e.g., puffed up fur for angry dog}.
}
		\label{fig:teaser}
	\end{center}%
}]

\maketitle
\def\thefootnote{*}\footnotetext{These authors contributed equally to this work.}\def\thefootnote{\arabic{footnote}}
\ificcvfinal\thispagestyle{empty}\fi

\begin{abstract}
This paper proposes a method for generating images of customized objects specified by users. The method is based on a general framework that bypasses the lengthy optimization required by previous approaches, which often employ a per-object optimization paradigm.
Our framework adopts an encoder to capture high-level identifiable semantics of objects, producing an object-specific embedding with only a single feed-forward pass. The acquired object embedding is then passed to a text-to-image synthesis model for subsequent generation. 
To effectively blend a object-aware embedding space into a well developed text-to-image model under the same generation context, we investigate different network designs and training strategies, and propose a simple yet effective regularized joint training scheme with an object identity preservation loss. Additionally, we propose a caption generation scheme that become a critical piece in fostering object specific embedding faithfully reflected into the generation process, while keeping control and editing abilities.
Once trained, the network is able to produce diverse content and styles, conditioned on both texts and objects. We demonstrate through experiments that our proposed method is able to synthesize images with compelling output quality, appearance diversity, and object fidelity, without the need of test-time optimization. Systematic studies are also conducted to analyze our models, providing insights for future work.
\end{abstract}

\section{Introduction}
\label{sec:intro}
\input{1_intro}

\section{Related Work}
\label{sec:related_work}
\input{2_related}

\section{Approach}
\label{sec:approach}
\input{3_approach}

\section{Experiments}
\label{sec:experiments}
\input{4_experiments}

\section{Ablation Studies}
\input{5_ablation}

\section{Limitation and Societal Impact}
\input{6_limitation}

\section{Conclusion}
\input{7_conclusion}

\clearpage
\newpage
{\small
\bibliographystyle{ieee_fullname}
\bibliography{egbib}
}

\end{document}

%% file: 1_intro.tex
Text-to-image synthesis~\cite{saharia2022photorealistic,ramesh2021zero,yu2022scaling,rombach2022high,sauer2023stylegan,chen2022re,chen2023subject} has gained increasing attention and experienced rapid development with recent advances of GANs~\cite{goodfellow2020generative,karras2019style,brock2018large,karras2018progressive,karras2020analyzing,karras2021alias} and diffusion models~\cite{ho2020denoising,song2020denoising}. With sophisticated designs and enormous amount of training data~\cite{schuhmann2022laion,radford2021learning}, images with unprecedented quality and diversity can be generated through conditioning on free-form texts provided by users.
Apart from generic objects, an intriguing question would be \textit{whether it is possible to synthesize images capturing an object specified by users}. Generating a particular object requires the understanding of its high-level concept, which is intricate, if not impossible, when the desired object is not contained in the training set. 
To remedy the domain gap, existing methods~\cite{gal2022image,ruiz2023dreambooth,brooks2023instructpix2pix,kumari2022customdiffusion,mokady2022null} generally adopt a fine-tuning paradigm, where a text prompt representing the object and a pre-trained synthesis model are jointly optimized using multiple images of the object provided by users. The optimized text prompt is then combined with natural language descriptions to generate outputs containing the objects with various content and styles. 

Training a model for each object is infeasible to scale up for practical uses. In particular, as object-specific fine-tuning is required, the aforementioned paradigm is unable to produce fast adaptation to arbitrary objects. The applicability of such methods is further limited by the intractable model storage cost, which increases with the number of objects considered. Furthermore, these methods usually require multiple images of the same object, which is not always available in reality.

The main focus of this paper is \textbf{\textit{learning a single general model that is able to compose a new scene around given objects yet without the need of per-object optimization, using as few as one image}}. This is an unexplored direction in spite of its wide applicability. 
In this work, we introduce a general framework as the first step towards this goal. Unlike existing works~\cite{ruiz2023dreambooth,kumari2022customdiffusion,mokady2022null} that discretely align a target object with a unique prompt through iterative optimization, we aim to continuously project object embedding and text-to-image generation embedding into an unified semantic space.

Despite being straightforward, it remains non-trivial 1) how to adapt text-to-image models to the object embeddings, and 2) how to construct training data for improving personalization while preserving editing capability in the pre-trained models.
Given the object embedding as an extra input, we insert additional attention modules to a text-to-image network. The augmented network is then fine-tuned to condition also on the object embedding. However, directly fine-tuning the network with the object embedding results in deteriorated editing capability. We therefore propose a \textit{regularized joint training scheme} with an \textit{cross-reference regularization} to maintain editability while making the model object-conditioning.
In addition, we introduce a caption generation scheme for augmenting the training dataset. This essentially increases data specificity, leading to performance gain in terms of output quality, appearance diversity, and object fidelity. 

With personalized text-to-image synthesis becoming an indispensable task in content creation, an efficient algorithm is of utmost importance to improve its applicability. In this work, we take the first step towards this direction and demonstrate how the optimization process can be bypassed without sacrificing performance. Our resultant model possesses simplicity and produces a personalized image in a single feed-forward pass, leading to reduced computational and storage costs.

%% file: 2_related.tex
\noindent\textbf{Text-to-Image Synthesis.}
Approaches for text-to-image synthesis~\cite{saharia2022photorealistic,ramesh2021zero,yu2022scaling,rombach2022high,sauer2023stylegan,kawar2023imagic,chang2023muse,ramesh2021zero,ramesh2022hierarchical,yu2022scaling,pan2023arbitrary,sheynin2022knn,nichol2022glide,nichol2022glide,wu2022adma,liao2022text} can be divided into three main categories. 
The vector-quantized approach~\cite{yu2022scaling,chang2023muse,ding2022cogview2} first learns a discrete codebook through training an autoencoder. After training, earlier works~\cite{yu2022scaling,ding2022cogview2} adopt an autoregressive transformer to predict the tokens sequentially. The predicted tokens are then passed to the decoder to generate the output image. To reduce computational cost for high-resolution images, bidirectional transformers~\cite{chang2023muse} are introduced to predict the tokens all at once and iteratively refine them. In this case, the computational time is reduced by eliminating the sequential operations.
Diffusion models~\cite{rombach2022high,saharia2022photorealistic,sheynin2022knn,nichol2022glide} synthesize images through iterative denoising. Starting from a standard Gaussian noise, a UNet is usually adopted to denoise the intermediate outputs, conditioning on the text prompt, to produce a less noisy outputs. The final output is obtained through repeating the denoising process. 
Recently, StyleGAN-T~\cite{sauer2023stylegan} demonstrates the capability of the GAN framework~\cite{tao2022df,huang2022dse,liao2022text} in text-to-image synthesis by modifying the StyleGAN~\cite{karras2018progressive,karras2020analyzing,karras2021alias} architecture, allowing text conditioning.
The aforementioned approaches achieve compelling performance with the presence of large-scale text-image datasets~\cite{schuhmann2022laion}. In this work, we follow the diffusion model paradigm. In particular, we adopt Imagen~\cite{saharia2022photorealistic} as our backbone network.

\begin{figure*}[t]
    \centering
    \includegraphics[width=0.99\textwidth]{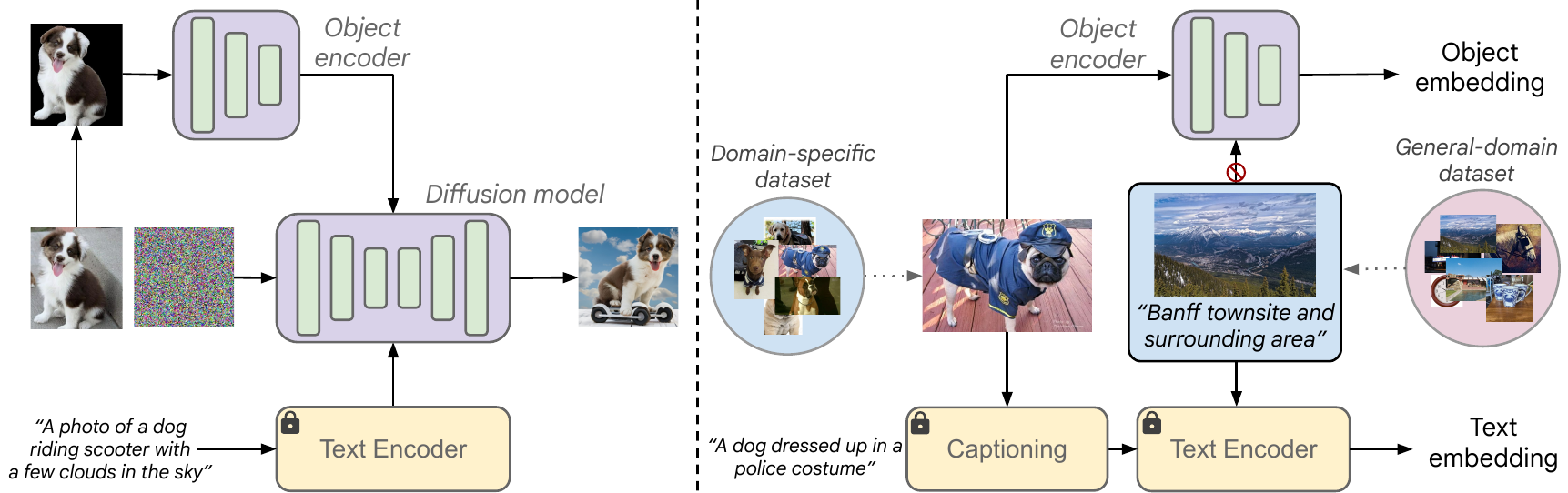}
    \caption{
        \textbf{Framework Overview.} \textbf{(Left)} Given a reference image, the background is removed, and an encoder is used extract embedding from the filtered image. The object embedding is then used along with the text embedding for subsequent generation. \textbf{(Right)} Our triplet preparation scheme. For domain-specific dataset, which generally does not caption, we apply a captioning model to obtain corresponding caption. Note that the object embedding is set to a null embedding for general-domain images. 
    }
    \label{fig:overview}
\end{figure*}

\vspace{0.1cm}
\noindent\textbf{Personalized Synthesis.}
Most existing works for personalized synthesis adopt a pre-trained synthesis network and perform test-time training for each object. For instance, MyStyle~\cite{nitzan2022mystyle} adopts a pre-trained StyleGAN for personalized face generation. For each identity, it optimizes the latent code as well as fine-tuning the pre-trained network. Once trained, it can be used to synthesize images of the target identity.
This paradigm is also seen in the task of text-to-image synthesis~\cite{gal2022image,ruiz2023dreambooth,brooks2023instructpix2pix,kumari2022customdiffusion,mokady2022null}. Given a pre-trained text-to-image synthesis model and multiple images containing the target object, a text prompt representing the objects are optimized and the network is optionally fine-tuned to further adapt to the target object. After training, the optimized object text prompt can be combined with natural language descriptions to generate diverse outputs.
With test-time optimization required, the aforementioned approach usually requires minutes to hours for each object. In addition, the model storage cost increases with the number of objects to be handled. As a result, their scalability and practicality are essentially limited. In this work, we focus on bypassing the lengthy optimization, enhancing the efficiency of personalized text-to-image synthesis.

%% file: 3_approach.tex
\subsection{Overview}
\label{subsec:overview}
\noindent\textbf{Framework.}
As shown in Fig.~\ref{fig:overview}, our framework consists of two main components. The first component is a text-to-image synthesis network~\cite{saharia2022photorealistic,rombach2022high}. We use a pre-trained model to enable text-conditioning synthesis. The second component is an image encoder. It is used to encode arbitrary objects into object embeddings, which serve as the second condition of the synthesis network for object-specific generation. To accommodate the object embedding which is not part of the pre-trained model, we augment the network by inserting cross-attention modules.

Specifically, let $\mathbf{x}$ be an image containing the object of interest and $\mathbf{c}$ be text caption describing a desired output image. The image object encoder $\mathcal{I}$ and text encoder $\mathcal{T}$ are used to compute the object embedding $\mathcal{I}(\mathbf{x})$ and text embedding $\mathcal{T}(\mathbf{c})$, respectively. The embeddings are then passed to the augmented text-to-image diffusion model, such as Stable Diffusion~\cite{rombach2022high} and Imagen~\cite{saharia2022photorealistic}, to generate the final output. It is noteworthy that our framework is generic and is not confined to a particular architecture. In this work, we adopt Imagen~\cite{saharia2022photorealistic} as the synthesis network.

Unlike most existing works that perform iterative optimization for each object, our proposed framework only require a single forward pass to generate images of an object. This essentially reduces the computation and storage overhead incurred by per-object optimization.

\vspace{0.1cm}
\noindent\textbf{Encoder.}
The object encoder $\mathcal{I}$ is responsible for capturing the concept of an object and is essential in our framework. In theory, one can use networks of any forms as the encoder, but we observe significant performance differences with different choices of networks. Our hypothesis is that in order to capture abstract concept, the encoder should be trained with 1) large amount of data and 2) an objective relating concrete objects with abstract description. Our hypothesis is corroborated by our studies in Fig.~\ref{fig:encoder}, where CLIP~\cite{radford2021learning} outperforms VGG~\cite{simonyan2015very}. Therefore, we adopt a frozen pretrained CLIP image encoder for simplicity in this work, but it worth mentioning that one can use a more sophisticated encoder for potential performance gain. The text encoder $\mathcal{T}$ is a frozen T5-XXL~\cite{raffel2020exploring} as in Imagen. 

\subsection{Preliminary}
\label{subsec:background}
\noindent\textbf{Imagen~\cite{saharia2022photorealistic}.}
Conditioned on text embeddings $\mathbf{c}$, Imagen is trained with a denoising objective
\begin{equation}
\label{eq:ddpm}
    \mathbb{E}_{\mathbf{x}, \mathbf{c}, \epsilon, t}\left[||\epsilon_\theta(\mathbf{x}_t, t, \mathbf{c}) - \mathbf{\epsilon}||_2^2\right],
\end{equation}
where $\mathbf{x}_t$ is a noisy version of the groundtruth image, $\epsilon\,{\sim}\,\mathcal{N}(\mathbf{0}, \mathbf{I})$ denotes standard Gaussian, and $t$ is the timestep. 

Classifier-free guidance~\cite{ho2022classifier} is a commonly used technique for improving diffusion model sample quality. It efficiently train one model for both conditional and unconditional objectives, achieved by randomly dropping the condition $\mathbf{c}$. During sampling, the intermediate predictions are adjusted based on the conditional and unconditional outputs as:
\begin{equation}
    \hat{\epsilon}_\theta(\mathbf{x}_t, \mathbf{c}) = w\epsilon_\theta(\mathbf{x}_t, \mathbf{c}) + (1 - w)\boldsymbol\epsilon_\theta(\mathbf{x}_t, \mathbf{c}_\emptyset),
\end{equation}
where $w$ is the guidance weight and $\mathbf{c}_\emptyset$ is the embedding of an empty text string. Intuitively, the guidance effect increases with $w$.

Instead of directly synthesizing images at the target resolution of $1024{\times}1024$, Imagen adopts a cascaded structure. It first generates an $64{\times}64$ image and uses two text-conditioning super-resolution models to upsample the image to $256{\times}256$ and then to $1024{\times}1024$.

\subsection{Triplet Preparation}
\label{subsec:triplet}
This section discusses the keys in preparing triplets for our network fine-tuning. Specifically, we first discuss our caption generation scheme, followed by our object embedding generation.

\vspace{0.1cm}
\noindent\textbf{Captioning with PaLI~\cite{chen2022pali}.}
One could improve personalization by training with objects of the same category. For example, the identity preservation of a dog is improved when datasets containing animals are included during training. However, such datasets usually contain no text captions, prohibiting domain-specific fine-tuning. 

The most intuitive way to generate captions is to apply a classification network to the images and set the caption to the class name. This method is simple to operate, but is unable to generate descriptive captions. This essentially leads to reduced capability in text-conditioning.

In this work, we propose to apply a language-image model, PaLI~\cite{chen2022pali}, on the images to generate descriptive captions. Let $f_c$ be a captioning model, we apply it to the image $\bf{x}$ to obtain the \textit{coarse caption} $\mathbf{c}_c\,{=}\,f_c(\mathbf{x})$.
In addition to the general description provided by PaLI, we further incorporate concrete attributes (\eg,~face attributes) if they are available. Let $f_a$ be the attribute classification network, our \textit{fine caption} $\mathbf{c}_f\,{=}\,f_a(\mathbf{x})$ is generated by applying $f_a$ to the image.
The two captions are then concatenated as the final caption $\mathbf{c}$.
Our approach generates captions with both abstract and concrete description, circumventing the loss of text-conditioning property.

\vspace{0.1cm}
\noindent\textbf{Background-Masked Object Embedding.}
Intuitively, the object embedding should depend only on the object, and should be agnostic to the background. Therefore, to better disentangle the object from the input image, we apply a binary mask to remove the background. Let $f_m$ be the mask generation function, we have $\mathbf{o} = f_o(\mathbf{x}\otimes\mathbf{m})$,
where $\otimes$ denotes pointwise multiplication, and $\mathbf{m}\,{=}\,f_m(\mathbf{x})$ denotes the binary mask. In such a way, the object embedding depends only on the object, possessing higher object specificity.

\subsection{Regularized Joint Training}
\label{subsec:training}
Overfitting a pretrained model with only few images sometimes can lead to a potential danger that the model forget precious skills gained during pretraining~\cite{he2019analyzing}. However our framework target a large collection of objects from similar categories, which makes joint training with conventional text-to-image data possible. Moreover, Personalization could be improved through training on large-scale text-image dataset and domain-specific dataset jointly. However, directly mixing the two datasets results in the loss of text-conditioning capability with the object embedding dominating the generation. It remains a challenging task to effectively condition the network on the extra object embedding without compromising the text-to-image synthesis capability. We propose a \textit{regularized joint training scheme} to 1) avoid dominance of the object embedding and 2) enhance object fidelity. 

\begin{figure}[t]
    \centering
    \includegraphics[width=0.49\textwidth]{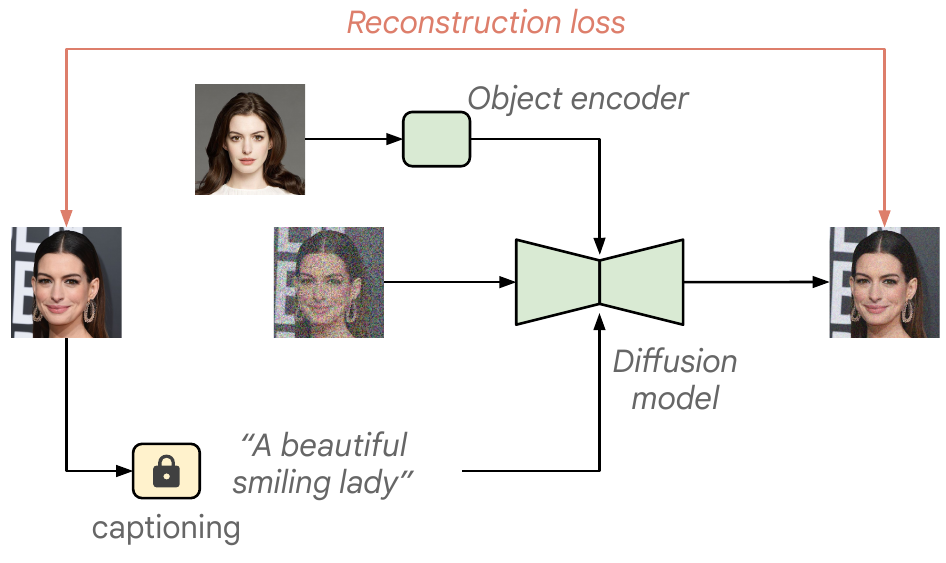}
    \caption{
        \textbf{Cross-reference regularization.} Since object embedding is agnostic to appearances, we use another image of the same identity to compute object embedding, encouraging disentanglement of identity information.
    }
    \label{fig:preservation_loss}
\end{figure}

\begin{figure*}[t]
    \centering
    \includegraphics[width=0.99\textwidth]{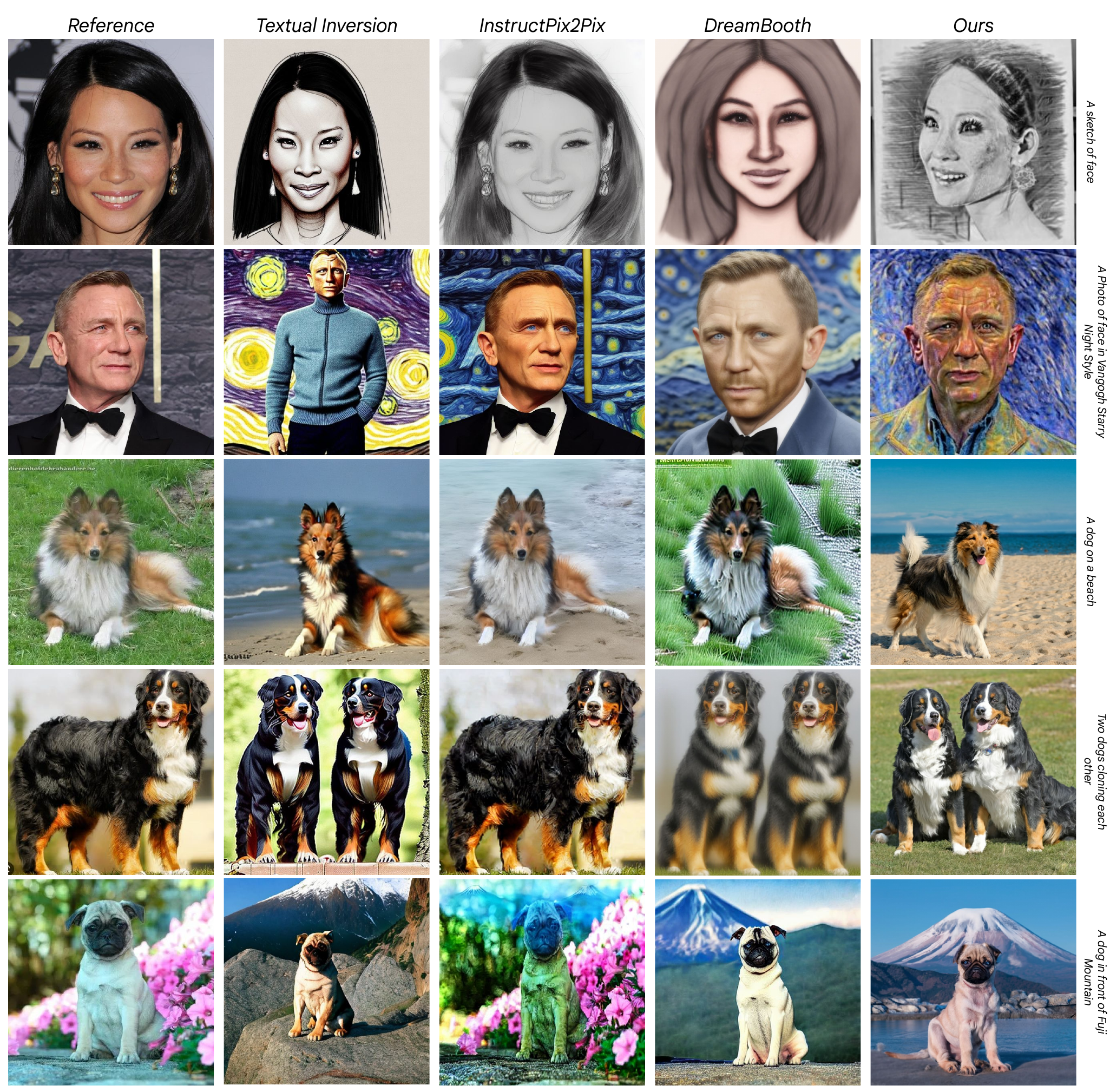}
    \caption{
        \textbf{Qualitative Comparison.} For human face styling, existing works sometimes either struggle to generate styles faithful to the texts or fail to preserve identity. For animals, existing works often overfit to the input image, generating outputs with limited diversity. Note that for Textual Inversion and DreamBooth, five images are used as input for human faces and one image is used for animals. 
    }
    \label{fig:result}
\end{figure*}

\vspace{0.1cm}
\noindent\textbf{Cross-reference regularization.}
Motivated by the fact that the object embedding is shared across images of the same object, we propose the cross-reference regularization to encourage disentanglement of identity information from the object embedding (Fig.~\ref{fig:preservation_loss}). Given an additional image $\mathbf{\bar{x}}$ capturing the same object, we randomly replace the object embedding $\mathbf{\bar{o}}$ by that computed from $\mathbf{\bar{x}}$:
\begin{equation}
    \mathbf{z} = 
\begin{cases}
(\mathbf{x}, \mathbf{\bar{o}}, \mathbf{t})            & \text{if }p<\omega,\\
(\mathbf{x}, \mathbf{o}, \mathbf{t})     & \text{otherwise},\\
\end{cases}
\end{equation}
where $p\,{\sim}\,\text{Uni}(0, 1)$ and $\omega$ is a pre-determined threshold. Here $\mathbf{z}$ represents the training tripelet. Note that the regularization is applied only for domain-specific images. Then, the synthesis network learns to distill object-specific information from the object embedding, essentially removing image-specific clues. We find that this significantly improve the object fidelity.

\vspace{0.1cm}
\noindent\textbf{Object-Embedding Dropping.}
In practice, one could apply the object encoder to the images in the general-domain dataset. However, we observe that the generation process is dominated by the object embedding. As a result, the trained network generates outputs solely based on the object embedding, neglecting the text conditions. To avoid this, we impose an implicit regularization to reduce the reliance on the object embedding, especially for general-domain images. Let $\mathbf{z}$ be the triplet during training, the object embedding is set to a null embedding for general domain images:
\begin{equation}
    \mathbf{z} = 
\begin{cases}
(\mathbf{x}, \mathbf{o}, \mathbf{t})            & \text{if } \mathbf{x}\text{ is domain-specific},\\
(\mathbf{x}, \boldsymbol{\phi}, \mathbf{t})     & \text{otherwise},\\
\end{cases}
\end{equation}
where $\boldsymbol{\phi}$ denotes a fixed null embedding. In this case, the network learns to leverage object embedding for domain-specific objects, while keeping the capability of text-conditioning through reconstructing general-domain images solely with text embedding.

\vspace{0.1cm}
\noindent\textbf{Whole-Network Tuning.}
In the test-time optimization paradigm, a recent work~\cite{kumari2022customdiffusion} discovers that fine-tuning only the attention modules lead to a comparable results. However, it is observed that training only the extra attention module while fixing the pre-trained backbone results in inferior performance in our framework. In particular, when training only the attention layer, the network is inferior in preserving the identity of the object despite being able to synthesizes corresponding context based on the text captions. Therefore, we unfreeze all the network weights and train them jointly.

%% file: 4_experiments.tex
\noindent\textbf{Settings.}
We adopt pre-trained Imagen~\cite{saharia2022photorealistic} as the text-to-image network. The text embeddings are computed by T5-XXL~\cite{raffel2020exploring}. We use CLIP~\cite{radford2021learning} image encoder to generate the object embeddings, and insert additional attention modules to Imagen to accommodate the object embeddings. The entire network except the object encoder is jointly trained with the same objective as used in training Imagen.
We evaluate our method on two categories. For human faces, we incorporate the CelebA~\cite{liu2015deep} into our internal large-scale text-image dataset and train jointly on the combined datasets. For animals, we adopt LSUN-dog~\cite{yu2015lsun} for fine-tuning. 
The models are fine-tuned for around two days with 64 TPU-v4 chips. More details on the architecture and training settings are discussed in the supplementary material.
We compare our performance with three existing state of the arts, namely Textual Inversion~\cite{gal2022image}, DreamBooth~\cite{ruiz2023dreambooth}, and InstructPix2Pix~\cite{brooks2023instructpix2pix}. We use their publicly released code, for Dreambooth, we use the implementation in diffuser for comparison. 

\noindent\textbf{Comparison.}
Fig.~\ref{fig:result} shows the comparison between our method and existing works. We first demonstrate style editability of our method on human faces. Existing works sometimes either struggle to generate styles faithful to the texts or fail to preserve identity details. For example, all methods except ours are unable to produce a sketch-like image given the text ``A sketch of face''. It is worth noting that Textual Inversion and DreamBooth take five image as input, while our method requires only one.   

We then demonstrate our capability on synthesizing diverse context with animals. Given only one image as input, we see that existing works often overfit to the input image in terms of pose and gesture, generating outputs with limited diversity. 
In contrast, our method robust to the number of input images and is able to generate images with diverse poses and context. For example, only our method is able to generate two identical dogs with different poses. Furthermore, although our method is not trained on cat-specific and horse-specific datasets, it is generalizable to a wider class (\ie,~animals), as shown in Fig.~\ref{fig:teaser}.

In addition to generalizability and quality, our method possesses greater efficiency. Specifically, while the training and storage costs of the methods in comparison increases linearly with the number of objects, our method does not require any per-object training, and hence the costs remain constant. The aforementioned strengths of our method essentially ease the use of personalized synthesis, unleashing human creativity.

We further compare with baselines using quantitative metrics, including object similarity (OSim.), caption similarity (TSim.) and Kernel Inception Distance (KID)~\cite{binkowski2018demystifying}. Object similarity measures the distance between input subject and personalized output using the pretrained CLIP image encoder. Caption similarity measures the distance between personalized output and the prompt using the pretrained CLIP text encoder. KID is useful in evaluating how close our personalized output distribution is to the style given by prompts. Table~\ref{tab:quantitative_cmp} consistently demonstrates that the proposed approach performs better in preserving object identity and matching user prompts with a single input.
\begin{table}[!t]
    \centering
    \caption{Quantitative comparison. (TI-$n$: Textual Inversion with $n$ image. DB: DreamBooth. Pix2Pix: InstructPix2Pix.)}
    \scalebox{0.9}{
    \begin{tabular}{l|cccccc}
    \toprule
        Methods & TI-1 & TI-5 & DB-1 & DB-5 & Pix2Pix & Ours \\ \midrule
        OSim.$\uparrow$ &  0.23 & 0.31 & 0.33 & 0.34 & 0.37 & \textbf{0.46} \\
        TSim.$\uparrow$ & 0.18  & 0.20 & 0.28 &  0.31 & 0.36 & \textbf{0.35} \\
        KID$\downarrow$ & 20.34 & 17.31 &  24.57 & 16.89 & 15.08 & \textbf{13.23} \\ \bottomrule
    \end{tabular}
    }
    \label{tab:quantitative_cmp}
\end{table}

\begin{figure}[t]
    \centering
    \includegraphics[width=0.49\textwidth]{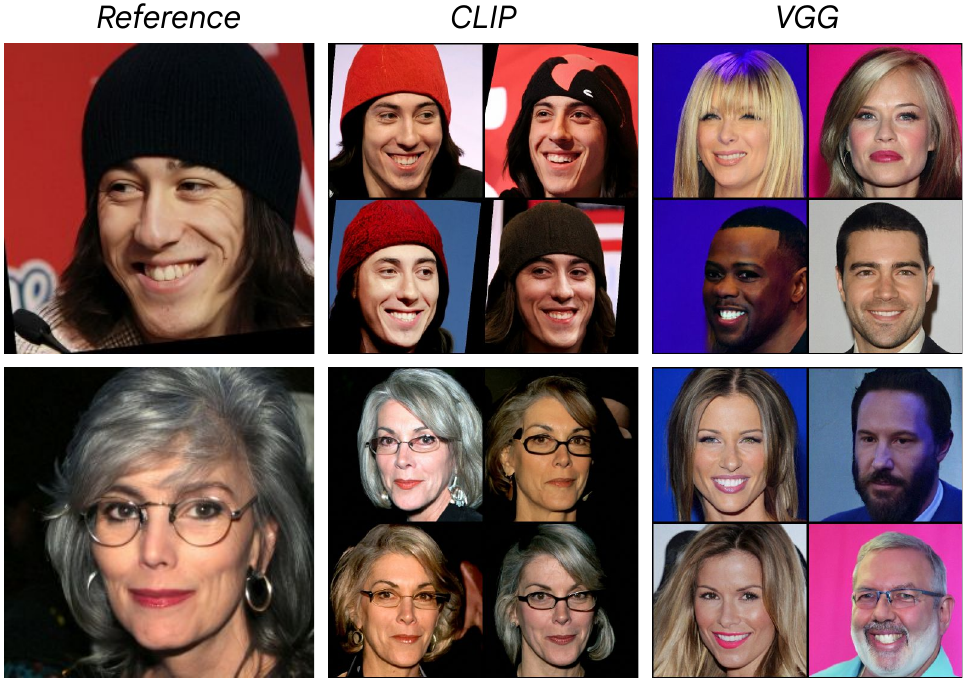}
    \caption{
        \textbf{Choice of Encoder.} The CLIP image encoder preserves identity while allowing appearance variations. In contrast, VGG generates random faces. 
    }
    \label{fig:encoder}
\end{figure}

%% file: 5_ablation.tex
\begin{figure}[t]
    \centering
    \includegraphics[width=0.49\textwidth]{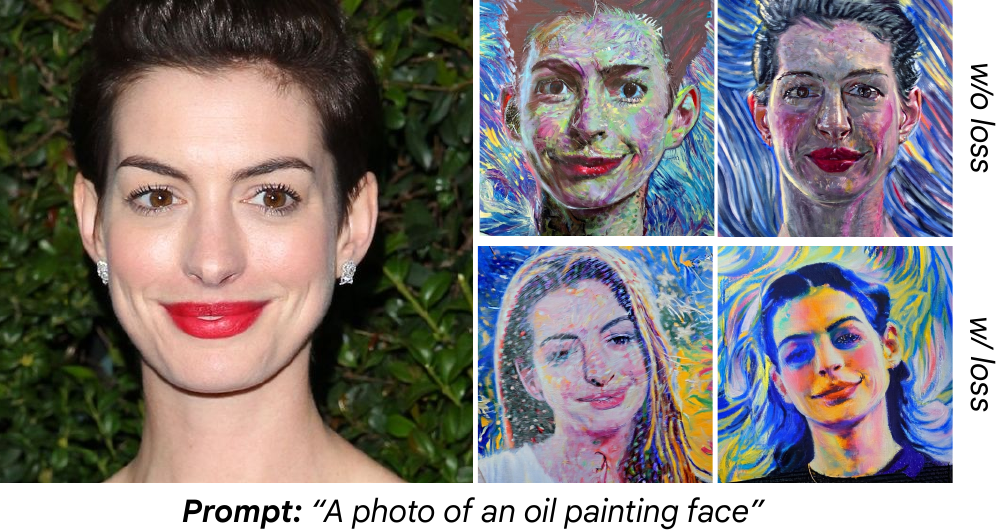}
    \caption{
        \textbf{Cross-reference regularization.} Our loss disentangles object concept from image-specific clues, leading to outputs with better \textit{identity preservation} and \textit{appearance diversity}. 
    }
    \label{fig:loss}
\end{figure}

\noindent\textbf{Choice of Encoder.}
We train a diffusion model on faces with object embedding as the sole condition to demonstrate the importance of choosing an appropriate encoder. 
As depicted in Fig.~\ref{fig:encoder}, when using VGG19~\cite{simonyan2015very}, which is trained for classification, as the encoder, the object embedding is unable to capture high-level concept of a face, thus generating random faces. 
In contrast, when CLIP is adopted, the network is able to generate faces with the same identity. Moreover, through relating abstract concept and concrete objects during the training of CLIP, the object embedding is agnostic to image-specific clues, leading to outputs with variations. This demonstrates the importance of the encoder. More sophisticated designs of the encoder are left as our future work.

\vspace{0.1cm}
\noindent\textbf{Cross-reference regularization.}
Since CLIP is not dedicated for identity preservation, we observe that directly using CLIP embedding leads to imperfect identity preservation and excessive retention of image clues. From Fig.~\ref{fig:loss} we see that without regularization, the synthesized images often overfit to the fine details (\eg,~hair styles) or insufficiently capture the identity. 
Our proposed regularization scheme alleviates the issues by swapping the object embeddings from the same object, enforcing identity disentanglement. As a result, the generated images possess greater diversity while preserving the identity.

\begin{figure}[t]
    \centering
    \includegraphics[width=0.49\textwidth]{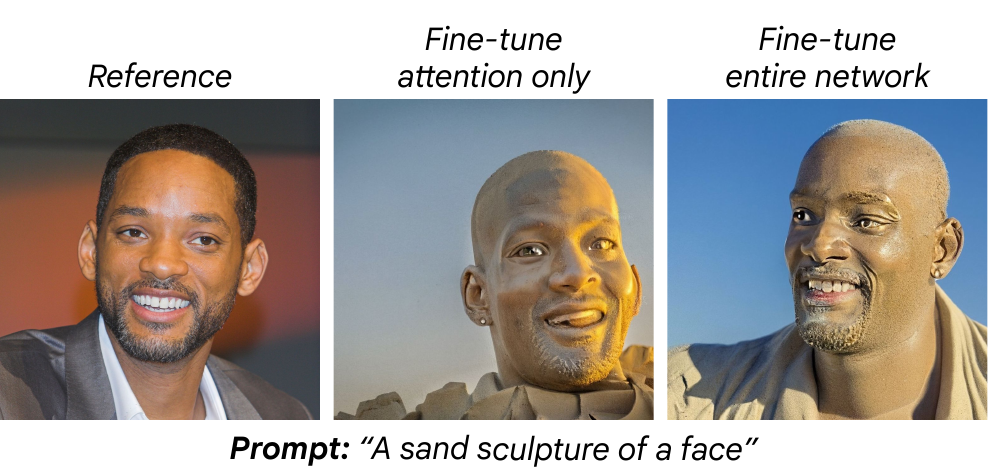}
    \caption{
        \textbf{Whole-Network Tuning.} Fine-tuning the entire network more effectively preserves the identity. 
    }
    \label{fig:tuning}
\end{figure}

\begin{figure}[t]
    \centering
    \includegraphics[width=0.49\textwidth]{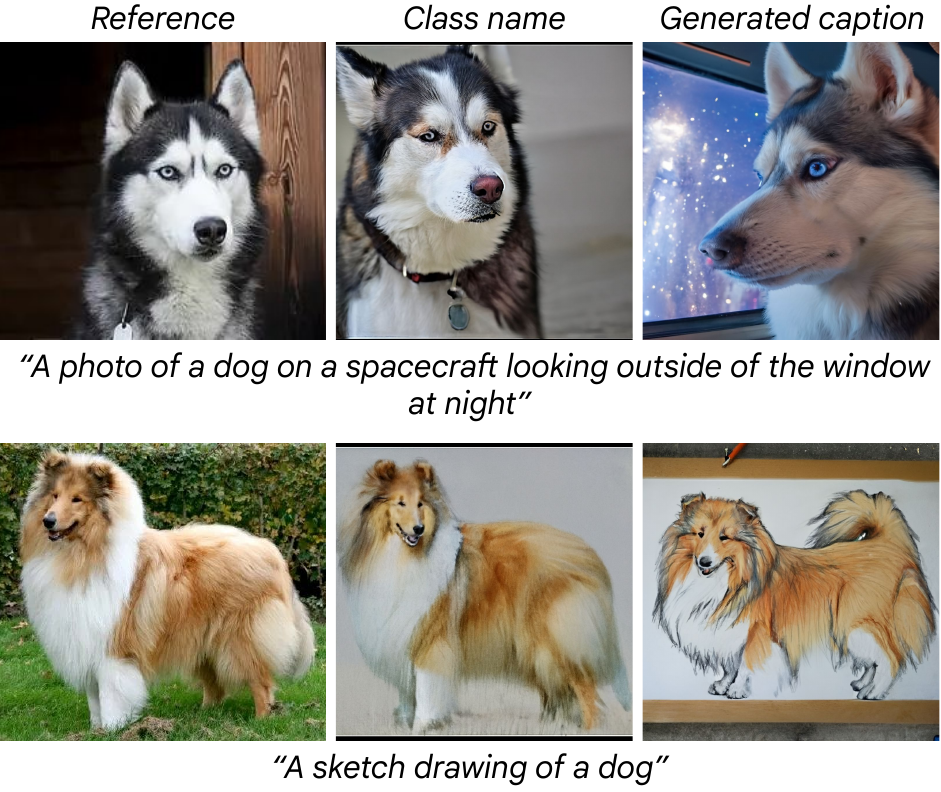}
    \caption{
        \textbf{Caption Generation.} Our caption generation scheme remedies the domain gap between the domain-specific dataset and general-domain dataset. The network trained with our generated captions possesses outputs faithful to the text input.
    }
    \label{fig:caption}
\end{figure}

\begin{figure*}[t]
    \centering
    \includegraphics[width=0.99\textwidth]{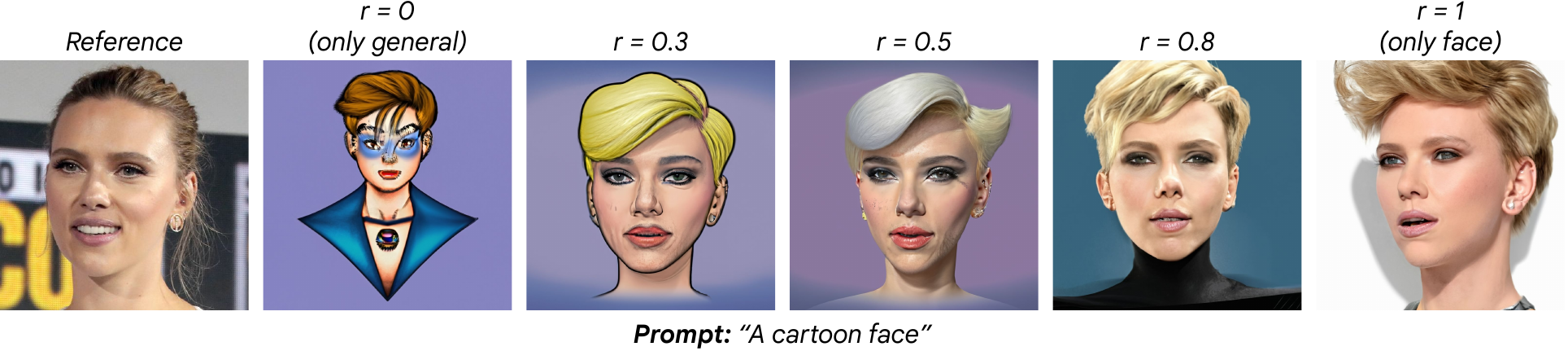}
    \caption{
        \textbf{Ratio of Dataset Mixing.} When trained only with the general-domain images ($r\,{=}\,0$), the network is unable to preserve identity. When trained only on domain-specific dataset ($r\,{=}\,1$), the network loses text-conditioning capability. With $r\,{=}\,0.3$, the network achieves a balance between editability and identity preservation. We choose $r\,{=}\,0.3$.
    }
    \vspace{-0.4cm}
    \label{fig:dataset_ratio}
\end{figure*}

\vspace{0.05cm}
\noindent\textbf{Whole-Network Tuning.}
For methods involving test-time optimization, it is shown that fine-tuning only the attention modules leads to improved training efficiency without compromising output quality~\cite{kumari2022customdiffusion}. However, we find in our framework that training only the added attention modules results in inferior performance. Specifically, as shown in Fig.~\ref{fig:tuning}, while fine-tuning only the attention layer retains the text-conditioning capability, identity cannot be preserved. In contrast, both output quality and identity preservation are improved when all weights are jointly trained. Our conjecture is that, unlike previous works that are confined to limited objects, our network is trained to generalized to unseen objects, and whole-network tuning enables more effective exploitation of the object embedding, improving generalizability.

\vspace{0.05cm}
\noindent\textbf{Caption Generation.}
Our caption generation scheme provides diverse captions on domain-specific datasets for our joint training scheme. When compared to the na\"ive approach of simply setting the class name (\eg,~dog) as the caption, our autocaptioning scheme leads to significantly better performance. For example, as shown in Fig.~\ref{fig:caption}, when only trained with class name, while the network is able to generate the same objects, it is inferior in generating content faithful to the text captions. This is due to the domain gap between the captions in the general-domain dataset and the domain-specific dataset. In contrast, our training scheme remedies the domain gap between the two datasets by synthesizing descriptive captions. As a result, the trained network is able to preserve object identity as well as generating contexts according to the given text caption. For example, in the first example, while the network trained with only class name is unable to generate outputs related to ``spacecraft'' and ``night'', our method generates faithful results with high fidelity to the text caption.

\vspace{0.05cm}
\noindent\textbf{Distribution Mixing.}
As discussed in the previous section, it is possible to incorporate domain-specific datasets into large-scale text-image datasets for improving personalization.  
As shown in Fig.~\ref{fig:dataset_ratio}, on the one hand, when we fine-tune the network on the general-domain dataset only (second row), the network is able to produce an output conforming to the text (\ie,~a cartoon face), but fails in resembling the identity. On the other hand, when trained only on domain-specific dataset, the network ignores the texts and produces a natural face that resembles the reference identity. The gradual transition shows that it is essential to balance the ratio of the two datasets. Throughout our experiments, $r\,{=}\,0.3$ is used.

%% file: 6_limitation.tex
\begin{figure}[t]
    \centering
    \includegraphics[width=0.47\textwidth]{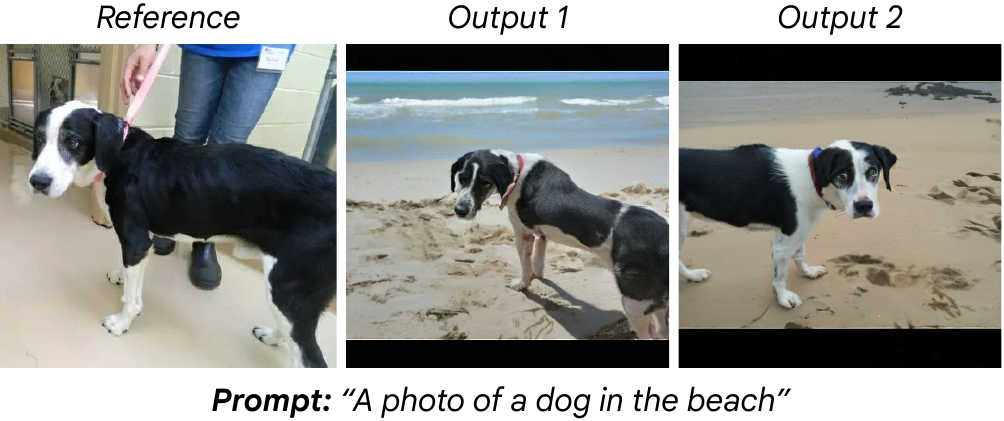}
    \vspace{-0.2cm}
    \caption{
        \textbf{Limitation.} Using only one image, the network could fail in generating the source's fine details. For example, the right eye of the dog disappears in output 1 and contains defects in output 2. \textbf{(Zoom in for best view)} 
    }
    \vspace{-0.3cm}
    \label{fig:limitation}
\end{figure}

We observe that the outputs of our method often contain defects when the corresponding details are not presented in the original image. For example, in Fig.~\ref{fig:limitation}, when the right eye of the dog is not shown in the input image, the network either ignores (output 1) or hallucinates (output 2) the right eye, and hence incoherence is observed in the outputs. In the future, we will extend the framework so that multiple images are taken as inputs, improving the robustness. 

This work can inherit the bias originated from training data, e.g, CelebA, which brings a consequent bias toward images of attractive people who are mostly in age range of twenty to forty years old. It may also contain only few images of certain group of race, which can potentially lead to misleading content creation and stereotyping propagation. Single image personalization may increase the ability to forge convincing images of non-public individuals. To prevent this, future efforts should be devoted to both the generative side (\eg,~cleaning training data) and discriminative side (\eg, forgery detection).

%% file: 7_conclusion.tex
This paper raises a question of whether the dominant approach of per-object optimization for personalized image synthesis is essential, and proposes a solution for the question.
We introduce a general framework of using an encoder to capture object concept so that test-time optimization can be bypassed. We then study the unique challenges in the framework. In particular, we propose a regularized joint training scheme to preserve object identity without compromising editing capability. We further propose an autocaptioning scheme to provide diverse text captions for better personalization. 
Our framework is able to synthesize images of the same object using texts provided by users using as few as one image in a single feed-forward pass, outperforming existing works in both quality and efficiency. We believe that the findings and insights in this work would inspire future works in improving the efficacy and applicability of personalized image synthesis approaches.